# SMPR: A structure-enhanced multimodal drug-disease prediction model for drug repositioning and cold start

Xin Dong[1,2,5], Rui Miao[3], Suyan Zhang[4], Shuaibing Jia[5], Leifeng Zhang[5], Yong Liang[6*], Jianhua Zhang[2,5*], Yi Zhun Zhu[2*]

1. Faculty of Innovation Engineering, Macau University of Science and Technology, 999078, Avenida Wai Long, Taipa, Macao, China
2. School of Pharmacy, Macau University of Science and Technology, 999078, Avenida Wai Long, Taipa, Macao, China
3. Basic Teaching Department, Zhuhai Campus of Zunyi Medical University, 519000, Zhu Hai, China
4. Faculty of Humanities and Arts, Macau University of Science and Technology, 999078, Avenida Wai Long, Taipa, Macao, China
5. Medical Engineering Technology and Data Mining Institute, Zhengzhou University, 450000, 100 Science Avenue, Zhengzhou, China
6. Peng Cheng Laboratory, 518055 Shenzhen, China

* Yong Liang, Jianhua Zhang and Yi Zhun Zhu are co-corresponding authors.

**Abstract:**
Repositioning drug-disease relationships has always been a hot field of research. However, actual cases of biologically validated drug relocation remain very limited, and existing models have not yet fully utilized the structural information of the drug. Furthermore, most repositioning models are only used to complete the relationship matrix, and their practicality is poor when dealing with drug cold start problems. This paper proposes a structure-enhanced multimodal relationship prediction model (SMRP). SMPR is based on the SMILE structure of the drug, using the Mol2VEC method to generate drug embedded representations, and learn disease embedded representations through heterogeneous network graph neural networks. Ultimately, a drug-disease relationship matrix is constructed. In addition, to reduce the difficulty of users' use, SMPR also provides a cold start interface based on structural similarity based on reposition results to simply and quickly predict drug-related diseases. The repositioning ability and cold start capability of the model are verified from multiple perspectives. While the AUC and ACUPR scores of repositioning reach 99% and 61% respectively, the AUC of cold start achieve 80%. In particular, the cold start Recall indicator can reach more than 70%, which means that SMPR is more sensitive to positive samples. Finally, case analysis is used to verify the practical value of the model and visual analysis directly demonstrates the improvement of the structure to the model. For quick use, we also provide local deployment of the model and package it into an executable program.

1. Introduction

The development of new drugs is a huge challenge of time and cost[1]. Using computer models to describe the complex relationships between drugs, diseases, and targets can accelerate the research process in terms of effectiveness, toxicity, and other aspects. Especially after the COVID-19 pandemic in 2019, methods such as drug repositioning to accelerate drug development have become even more important[2, 3]. Entecavir, Penciclovir, Ganciclovir, Indinavir, and other drugs have been reported to have potential effects in COVID-19[4]. However, the drugs that have been practically applied through drug repositioning are still limited, and a large number of drugs are still in clinical trials or abandoned at the stage III of clinical trials[5]. Therefore, supplementing information to further improve the accuracy of drug repositioning models remains a direction worthy of attention.

Multimodal data, which adds more prior knowledge to the prediction of drug-disease relationships, provides a new perspective for repositioning drugs from different data sources[6, 7]. The deepDR proposed by Zeng et al. integrates seven relational networks including drug, disease, side effects, and target networks to predict potential associations between drugs and diseases[8]. The NCH-DDA model utilizes node aggregation and neighborhood fusion to further enhance information exchange between different modalities[9]. Especially the REDDA model proposed by Gu et al., which utilizes five entities including drugs, proteins, genes, pathways, and diseases to construct heterogeneous networks, greatly enriches the knowledge that the model can learn[10].

However, most of the current multimodal data based repositioning models still lack sufficient consideration for the importance of drug structure, where compound structures are only used to construct network relationships between drugs. For the same protein active pocket, it is particularly important whether the drug can enter smoothly and has a group that can form a stable chemical bond with the residues therein[11, 12]. In addition, existing drug repositioning models are mostly used to complete input data sets, lacking application expansion in actual scenarios, which is a serious limitation on the generalization ability of the model[13]. Inputting new drugs for prediction with limited prior knowledge is called cold start (which do not exist in the training dataset), where the prediction of the relationship between unknown drugs and diseases is drug cold start[14].

To address the issues of repositioning mentioned above, we propose the SMPR model. The model includes a drug embedding module and a disease embedding module, which is an improvement on the existing model. The drug embedding module is inspired by NLP, treating the SMILE structure of compounds as a natural language and generating unique embedding representations for each compound using the Mol2vec model. The disease embedding modules utilizes heterogeneous networks to integrate multimodal information and jointly construct vector representations of diseases. Finally, the model reconstructs the drug disease relationship based on the features obtained from two modules. And in order to solve the cold start problem, we provide a simple cold drug start method based on the model repositioning learning results. For a drug that is not in the dataset, the drug embedding module is used to quickly obtain its representation, and based on the structural similarity weighting and prior knowledge of some existing diseases, the most likely disease to act is recommended.

2. Materials and Methods

2.1 Datasets

We collected a dataset used in our model, Dataset A (DA). DA is derived from the Fdataset[15], Cdataset[16], and KEGG[17] databases, containing 894 drugs, 454 diseases, and 2704 confirmed drug disease associations. More diverse knowledge of drugs, proteins, genes, pathways, and drug connectivity is sourced from DrugBank[18], CTD[19], STRING[20], and UniProt[21] databases. And as shown in Gu et al.[10], data preprocessing is carried out. In addition to the known drugs-proteins, diseases-genes, and drugs-diseases relationships, drugs-drugs and diseases-diseases matrix are constructed based on drug Extended Connectivity Fingerprint (ECFP) similarity and drug MeSH similarity. To validate the cold start capability of the model, we also divided the Dataset A into two parts: $Cold\_start_{train}$ (CS_train) and $Cold\_start_{test}$ (CS_test). DA is divided in a ratio of 9:1, where CS_train contains 805 drugs and CS_test contains 89 drugs. When validating the cold start later, the model will be retrained on CS_train and CS_test will be treated as a new drug that has not appeared in the training data.

Table 1. Introduction of Dataset A, CS_train and CS_test.

|  | Dataset A | $Cold\_start_{train}$ | $Cold\_start_{test}$ |
| --- | --- | --- | --- |
| Drug SMILE structures | 894 | 805 | 89 |
| Drugs- Drugs $R_{rr}$ | 799236 | 627653 | 0 |
| Drugs -Proteins $R_{rp}$ | 4397 | 3956 | 0 |
| Proteins-Genes $R_{pg}$ | 18545 | 18545 | 0 |
| Genes-Pathways $R_{gw}$ | 25995 | 25995 | 0 |
| Pathways-Diseases $R_{wd}$ | 19530 | 19530 | 0 |
| Diseases-Diseases $R_{dd}$ | 42500 | 42046 | 0 |
| Drugs-Diseases $R_{rd}$ | 2704 | 2425 | 279 |

In addition, to verify the stability of the model, we collected the data set used by Zhao et al. [22] to construct Dataset C (DC), in which the data only contains 579 drugs and 274 diseases, as shown in Table 2. We supplemented the structural information corresponding to the drug.

Table 2. Introduction of Dataset C.

|  | Dataset C |
| --- | --- |
| Drug SMILE structures | 579 |
| Drugs -Proteins $R_{rp}$ | 3219 |
| Diseases -Proteins $R_{dp}$ | 43131 |
| Drugs-Diseases $R_{rd}$ | 1897 |

2.2 Model Framework

The SMRP model, as shown in Figure 1, mainly includes disease embedding module and drug embedding module. In this chapter, we provide a detailed introduction to each module.

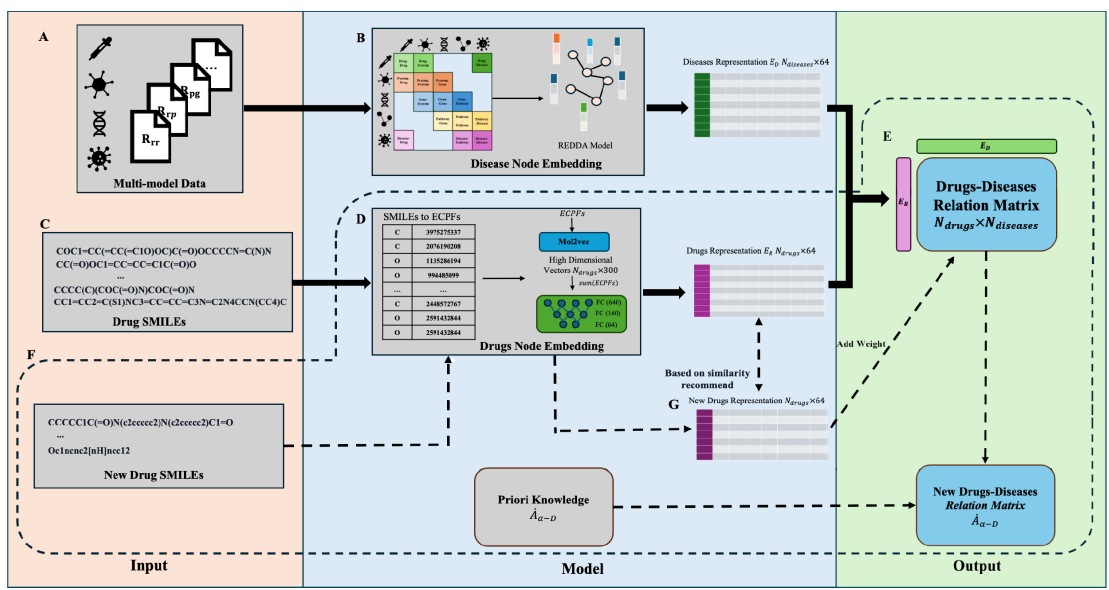

Figure 1. SMRP flowchart. The solid box represents the repositioning, and the dashed box represents the cold start. A. Multimodal data. B. The disease embedding module obtains disease embedding representations from heterogeneous networks through deep learning models. C. The input of SMILE structural data. D. The drug embedding module obtains a structure based embedding representation. E. Constructing a relationship matrix. F. New drugs of the cold start module. G. Weight the repositioning relationship matrix based on feature similarity.

2.2.1 Heterogeneous Network Construction and Node Feature Initialization

Firstly, a heterogeneous network G (N, E) is constructed based on multimodal data as the input for the disease embedding module (Figure 1A), where nodes include drugs, proteins, genes, pathways, and diseases. Based on the relationships organized in the dataset, a heterogeneous network is constructed, and the relationships are used as edge information in subsequent graph neural networks. Each node vector dimension is $N_{diseases} + N_{drugs}$. The features of drugs and diseases are assigned by their respective similarity matrices and the remaining node features are initially set as zero. The SMILE structure data of the drug is organized as input for the drug embedding module (Figure 1C).

2.2.2 Disease Embedding Module

The REDDA model is used as a disease embedding module. This model is a complex graph neural network with attention mechanism, which has received widespread attention for its excellent drug repositioning ability. The model first uses a fully connected network to initialize node features and map them to low dimensions. Then, two HeteroGCNs are used to globally learn the heterogeneous network, and one HeteroGCN sub network is used to learn each type of edge. Finally, one GAT module and one Layer Attention Block are used to obtain a 64 dimensional embedded representation of diseases in the heterogeneous network, denoted as $E_D$.

2.2.3 Drug Embedding Framework

In order to emphasize the consideration of drug structure in the model and solve the subsequent cold start problem, drug SMILE structure is used to generate embedded representations of drugs. Firstly, SMILE, as a concise structural feature, includes molecular connectivity information. We use ECFP to characterize the specific substructure of drug molecules[23]. Jaeger et al.[24], inspired by NLP, view

ECFP as a sentence and design the Mol2vec model to obtain embedded representations of substructures, improving the accuracy of molecular structure similarity retrieving. The model generate specific identifiers $ECPF_m$ for the substructures $m$ around each heavy atom in SMILE, and use the unsupervised method Mol2vec to convert $ECPF_m$ into a vector $v_m$. At the end of the module, the unbiased sum of all substructure vectors obtained from ECFP in a molecule is used to obtain the representation $V_R$ of the molecule. Firstly, the formula used the Mol2Vec model to generate the vector representation of substructure $m$ can be represented as:

$$v_m = Mol2Vec(ECPFm), m \in Substructures$$

Each drug embedding can be represented as:

$$V_R = \sum_{m \in Substructures} v_m$$

And according to Mol2vec's recommendation, 300 dimensional feature vector can maximize the reflection of molecular structural information. In order to embed $V_R$, three layers of fully connected network are used to encode the molecular representation and ultimately obtain a 64 dimensional drug embedding representation, denoted as $E_R$.

2.2.4 Building Drug Disease Relationship Matrix

For the embedding representations obtained from the disease embedding module and the drug embedding module, the model uses the product to obtain the drug disease correlation matrix $\hat{A}_{R-D}$, and normalizes it to a correlation score using the Sigmoid function:

$$\hat{A}_{R-D} = sigmoid(E_R E_D^T)$$

2.2.5 Drug Cold Start

When facing the cold start problem, the SMPR model is inspired by the feature similarity based recommendation system[25]. And our model's cold start is based on repositioning, providing users with a simple repositioning recommendation interface. When a new drug $\alpha$ that does not appear in the dataset, the model only calls the drug embedding module. The user inputs the SMILE structure of the drug, and the model compares its embedding representation with the saved drug embedding representation database to recommend the most likely diseases to the user (Figure 1.F). The model first uses the reciprocal of the Euclidean distance to calculate the correlation $\rho_\alpha$ between $E_\alpha$ and $E_R$:

$$\rho_\alpha = \frac{1}{\sqrt{(E_{\alpha 1} - E_{R1})^2 + \cdots (E_{\alpha 64} - E_{R64})^2}}$$

Subsequently, the repositioning result $\hat{A}_{R-D}$ is weighted using the correlation $\rho_\alpha$, and the relationship between the new drug α and the existing drug $A_{\alpha-D}$ is used as prior knowledge. The formula for the relationship $\dot{A}_{\alpha-D}$ is:

$$\dot{A}_{\alpha-D} = \rho_\alpha \times \hat{A}_{R-D} + A_{\alpha-D}$$

Then $\dot{A}_{\alpha-D}$ is normalized:

$$\dot{A}_{\alpha-D} = \frac{A_{\alpha-R} - \min(A_{\alpha-R})}{\max(A_{\alpha-R}) - \min(A_{\alpha-R})}$$

And recommend diseases for new drug based on the scores.

## 2.3 Model Parameter and Evaluation Indicators

For the stability of the SMPR model results, we divide the dataset using 10-fold cross validation and train 4000 epochs at a time. And the model removes the connection relationship of the test set in the graph during training. The model optimizer uses Adam, and the loss function uses a weighted BCEWithLogitsLoss function to balance the effects of different categories[26]. The formula can be denoted as:

$$Loss = \frac{N_{neg}}{N_{pos}} \text{BCEWithLogitsLoss}(Pred_{A_{D-R}} - Ground_{A_{D-R}})$$

where $N_{neg}$ and $N_{pos}$ are the number of relationships between labels 0 and 1 in the training set. In addition, the model learning rate is set to 0.005 and the dropout rate is 0.4. And we chose the AUC, AUPR、F1-Score, Accuracy, Recall, Specificity, and Precision proposed by Yu et al.[26] to evaluate our model.

In addition, to evaluate the potential impact of structural features on model performance, we used the t-SNE algorithm[27] to reduce feature dimensionality and the KMeans unsupervised clustering algorithm[28]. Since the drug-disease relationship matrix is a particularly discrete data, for the evaluation score of the disease relationship between drug $\gamma$ and drug $\delta$, we use the following formula:

$$R_{\gamma-\delta} = R_{\gamma-D} + R_{\delta-D}$$
$$score = N_{\gamma-\delta_{i=2}} / N_{\gamma-\delta_{i>1}}$$

## 3. Results

### 3.1 Comparative Experiment of Drug Repositioning Models

We ran repositioning task on the DA and compared the results with the models SMPR, REDDA, DRWBNCF[29], and DRAGNN[30]. To ensure the fairness of the comparison results, 10-fold cross validation was used for all models. The performance results are shown in Table 3. After paying more attention to structural information, the prediction performance of the SMPR model has indeed been improved, especially in the distinction of positive samples, with a recall score of 69%. For other evaluation indicators, AUC is consistently better than the comparison model in 10-fold cross validation, reaching 0.98 (Figure 2.). The AUPR index reaches 0.61, and the average value is the best in cross-validation. The F1-score and precision indicators of DRAGNN achieves the best results, while the scores of SMPR are 0.58 and 0.50 respectively.

Table 3. Model performance in DA.

| Method | AUC | AUPR | Accuracy | F1-score | Precision | Recall | Specificity |
|---|---|---|---|---|---|---|---|
| DRWBNCF | 0.8774 | 0.2961 | 0.9914 | 0.3682 | 0.3591 | 0.3778 | 0.9955 |
| DRAGNN | 0.9217 | 0.5500 | 0.9949 | 0.5696 | 0.6524 | 0.5055 | 0.9982 |
| REDDA | 0.9755 | 0.4042 | 0.9897 | 0.3778 | 0.3164 | 0.4686 | 0.9932 |
| SMPR | 0.9870 | 0.6106 | 0.9934 | 0.5825 | 0.5015 | 0.6949 | 0.9954 |

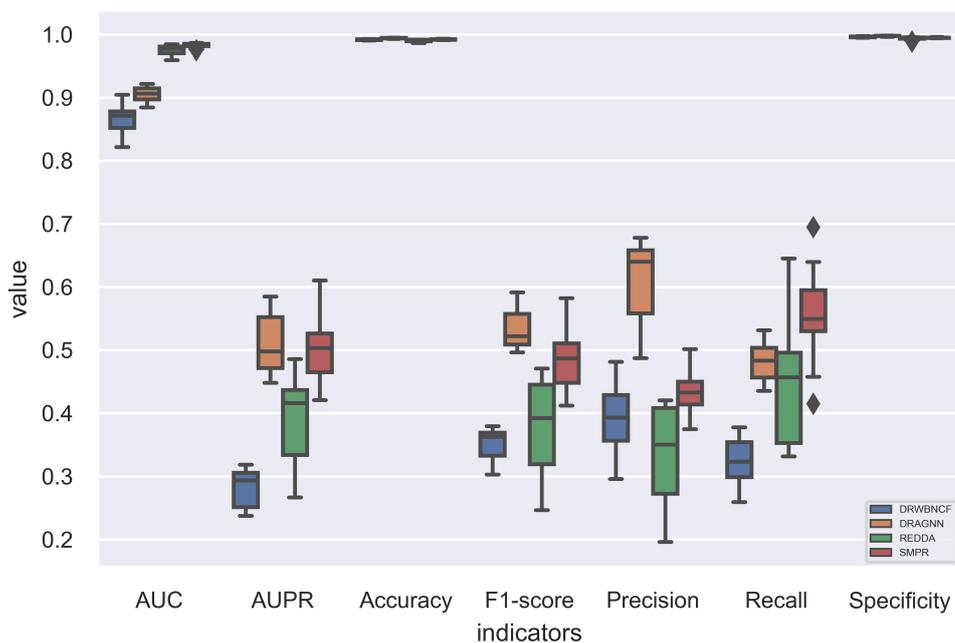

Figure 2. 10-fold cross validation results for 4 models.

3.2 Structural Feature Correlation Analysis

We combined the t-SNE dimensionality reduction algorithm to reduce the structure-based features obtained by the drug structure embedding module to 2 dimensions and put them into the Cartesian coordinate system for observation. In Figure 3.A, it can be roughly judged that it can be divided into 6 categories, and the KMeans unsupervised clustering method is used for clustering. The cluster boundaries in the figure are clear, and different categories can be clearly divided based on structural features. Finally, we sorted out the disease relationship in the corresponding data set of each type of drug and visualized the results using a heat map. In figure 3.C, the disease similarity of the drugs is calculated, and the red line distinguishes different categories. The value represents the average similarity between each category. It can be found that on the diagonal of the matrix, that is, the same type of drugs has the highest similarity in the diseases they act on. However, between different types, the similarity score of the disease they act on is significantly lower. This also proves that drugs with similar structures have similarities in their binding pocket, which will lead to potential similarities in the diseases they act on. Our model's enhanced focus on structure can effectively improve the model's prediction performance.

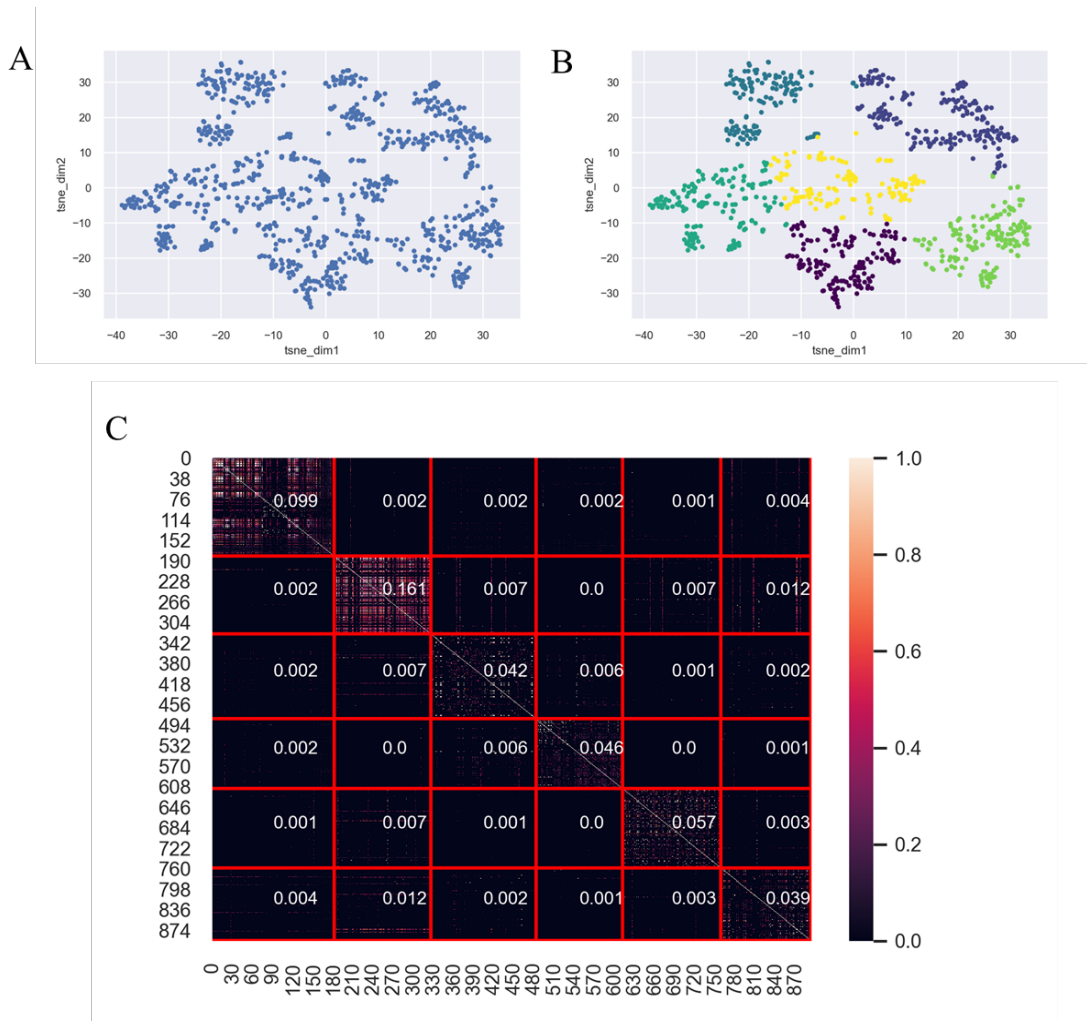

Figure 3. Analysis of the potential effects of structural features. A. Visualization results after t-SNE dimension reduction. B. Combining the KMeans to distinguish classes. C. Score the disease similarity of each drug-drug pair. The red line distinguishes different classes, and the value represents the average score of the comparison of different classes.

3.3 Sparse Matrix Tests

The SMPR model uses a variety of entities to enrich the connection relationships of heterogeneous networks. We hope that the model can alleviate the adverse effects of a large number of unknown relationships in this way. So we conducted sparse matrix tests in DA. As described in the existing study[31], the sparsity test removes approximately 20% of the drug-disease connections in the network. Here, to test the limits of our model's robustness, we remove $\varepsilon \in$ 错误!未定义书签。 of the edges. The results are shown in Figure 4. It can be found that the AUPR index is stable at 20%, 40%, and 60%, and is slightly lower than the complete data set. When the deleted edges reach 80%, the AUC, AUPR and Recall scores are 0.968, 0.369, and 0.424, respectively, which are still within an acceptable range. When all drug-disease connections are deleted, the model indicators drop significantly, and effective predictions can no longer be made. Therefore, for the 894 drugs, 454 diseases, and 2704 connections in DA, we only need to know 540 drug-disease relationships for the model to be effectively trained, which means that our model has strong robustness.

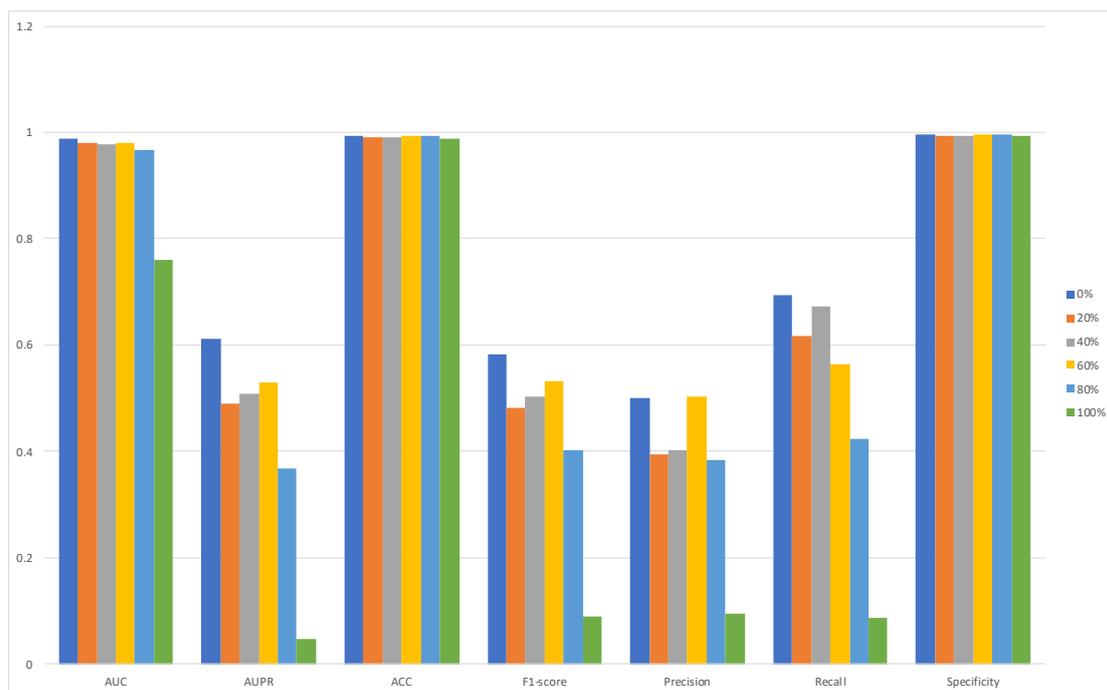

Figure 4. The SMPR model is tested on sparse matrices in DA. The edges with $\varepsilon \in \{0\%, 20\%, 40\%, 60\%, 80\%, 100\%\}$ are deleted.

3.4 Comparison Results of DC Dataset

We also verified the evaluation indicators of the model in the DC dataset to ensure the stability of the model in different public datasets. All four models passed the 10-fold cross validation. The modal information contained in DC is limited, including only drugs, proteins, and diseases. In addition, proteins do not have additional protein-protein interaction network (PPI) information. As shown in Figure 5, the SMPR model still shows the best prediction performance in AUC and AUPR. However, in the F1-score and Precision indicators, the DRAGNN model achieved the best results, while the REDDA and SMPR models performed poorly, which showed the same trend as the DA dataset. We consider that the SMPR model judges more negative samples as positive, which may be a limitation of the model framework. Richer information can help improve the distinction between positive and negative samples. In addition, the SMPR model performs particularly well in the Recall indicator, showing the model's excellent predictive ability for positive samples.

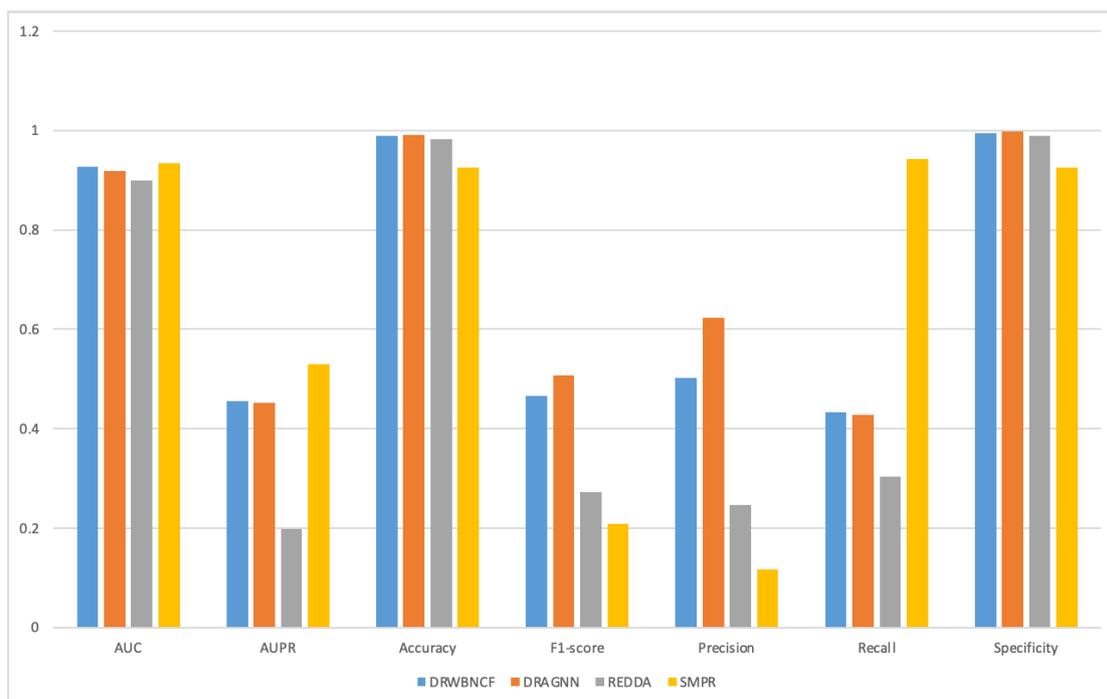

Figure 5. Performance of the four models on the DC dataset.

3.5 Case Study

We randomly selected two drugs from the SMPR repositioning results and conducted a literature review on the top 10 diseases that were most associated with their predictions. Baclofen (DB00181) is a drug used to relieve severe muscle spasms caused by certain diseases of the brain and spinal cord. The top 10 diseases in our recommended are mainly spasms and epilepsy. Most literature shows that Baclofen has a therapeutic effect on target diseases, while some diseases indicate that Baclofen can cause dysregulation of regulatory factors, leading to disease (Table 4). The relationship pathways between Baclofen and 10 diseases and their connections are shown in Figure 6. The five pathways with the highest connectivity are: hsa04921 (Oxytocin signaling pathway), hsa04723 (Retrograde endocannabinoid signaling), hsa04080 (Neuroactive ligand-receptor interaction), hsa04010 (MAPK signaling pathway), and hsa05412 (Arrhythmogenic right ventricular cardiomyopathy).

Another case, Docetaxel (DB01248), is an anti-cancer drug used to treat breast cancer, lung cancer, prostate cancer, head cancer, neck cancer and other cancers. Among the top 10 diseases recommended by the model, 9 are related to cancer, and literature shows that monotherapy or combination therapy can have therapeutic effects on the corresponding diseases (Table 5). In addition, cancer-related pathways have shown more complex regulatory mechanisms and richer connections. The pathways with the highest connectivity in Figure 7 are: hsa04144 (Ras signaling pathway), hsa05218 (Melanoma), hsa04210 (MAPK signaling pathway), hsa05220 (Transcriptional misregulation in cancer), and hsa04213 (Longevity regulating pathway).

Table 4. Literature survey of DB00181 and top 10 diseases.

| Drug | Disease | Description | Existing | Evidence |
|------|---------|-------------|----------|----------|

| DB00181 Baclofen | D010003 | Osteoarthritis | 0 | Abdelmonem et al.[32] |
| --- | --- | --- | --- | --- |
| | D006816 | Huntington Disease | 1 | Shoulson et al.[33] |
| | D013132 | Spinocerebellar Degenerations | 1 | Bushart et al.[34] |
| | D009128 | Muscle Spasticity | 1 | Pérez-Arredondo et al.[35] |
| | D000690 | Amyotrophic Lateral Sclerosis | 1 | Marquardt et al.[36] |
| | D015419 | Spastic Paraplegia, Hereditary | 1 | Margetis et al.[37] |
| | D004401 | Dysarthria | 0 | Leary et al.[38] |
| | D020190 | Myoclonic Epilepsy, Juvenile | 0 | Akgun et al.[39] |
| | D004832 | Epilepsy, Absence | 0 | Inaba et al.[40] |
| | D004829 | Epilepsy, Generalized | 0 | O.Carter Snead[41] |

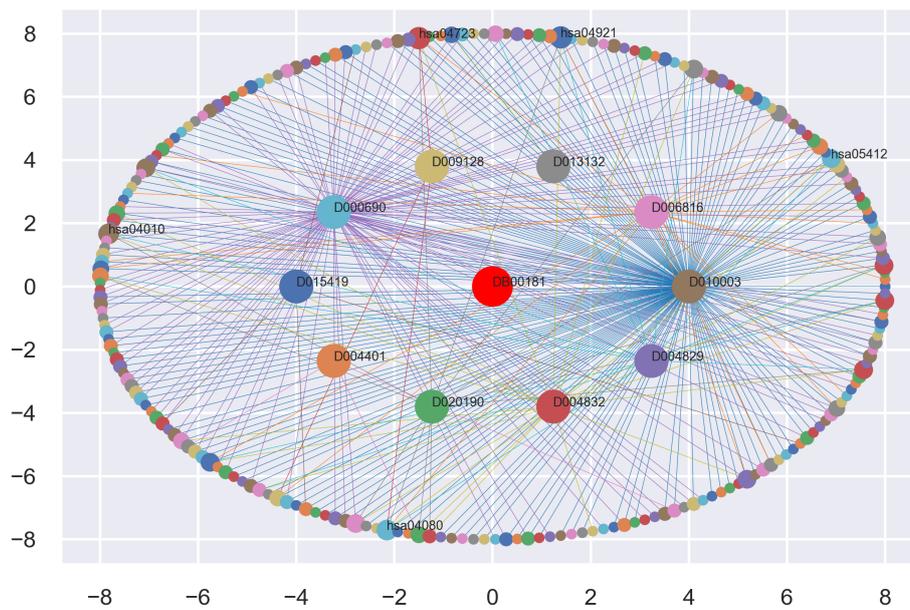

Figure 6. Intersection of pathways between DB00181 and the top 10 diseases. The top 5 pathways with the highest correlation are listed.

Table 5. Literature survey of DB01248 and top 10 diseases.

| Drug | Disease | Description | Existing | Evidence |
| --- | --- | --- | --- | --- |
| DB01248 Docetaxel | D015470 | Leukemia, Myeloid, Acute | 0 | Consolini et al.[42] |
| | D010190 | Pancreatic Neoplasms | 0 | Ryan et al.[43] |
| | D013274 | Stomach Neoplasms | 1 | Kazuhiro et al.[44] |
| | D001932 | Brain Neoplasms | 1 | Shaw et al.[45] |

| | D015179 | Colorectal Neoplasms | 0 | Wang et al.[46] |
| | D001749 | Urinary Bladder Neoplasms | 0 | McKiernan et al.[47] |
| | D009447 | Neuroblastoma | 0 | Francesco et al.[48] |
| | D011087 | Polycythemia Vera | 0 | Kunthur et al.[49] |
| | D002292 | Carcinoma, Renal Cell | 0 | Marur et al.[50] |
| | D016889 | Endometrial Neoplasms | 1 | Miyahara et al.[51] |

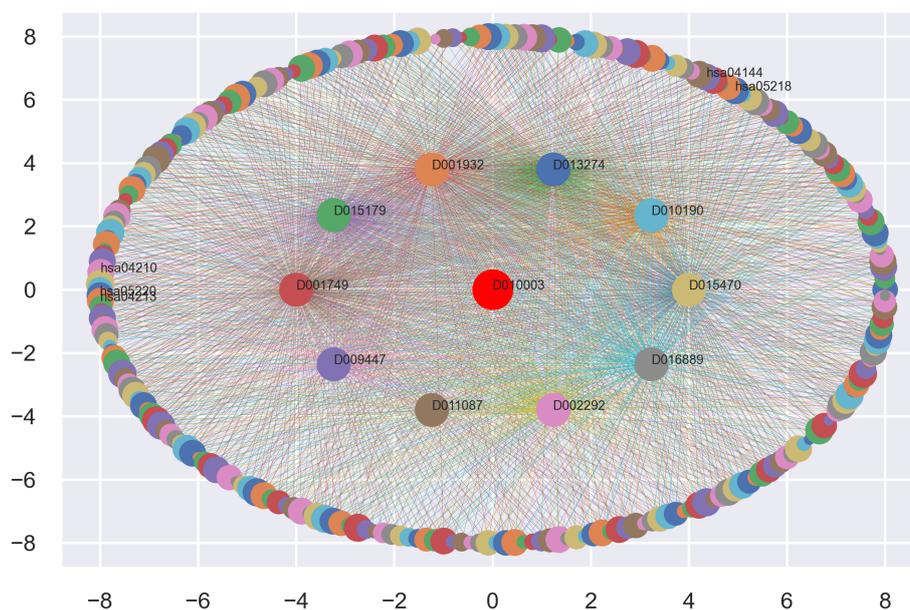

Figure 7. Pathway intersection between DB01248 and top 10 diseases. 10 diseases showed highly correlated pathways.

3.6 Results of Drug Cold Start

At the same time, our model focuses on the application of drug cold start. For a drug that is not in the dataset and we only know its SMILE structure, we also hope that the model can provide a simple interface. Therefore, we applied SMPR to simulate the cold start problem in CS_test. Firstly, we retrained the model on CS_train, and to ensure the reliability of the model, we used 5-fold cross validation. Then CS_test is used as a new dataset for cold start task testing.

According to the results shown in Table 6, when SMPR retrain on CS_train dataset, the results are stable, with AUC hovering around 0.97 and AUPR remaining around 0.45. And Figure 8 shows the cold start performance of CS_test. SMPR predicts new drugs for all diseases of CS_test based on similarity, the model believes that diseases with a score greater than 0.24 have potential effects based on the evaluation metrics mechanism. Meanwhile, the model shows a certain predictive ability, especially in the division of positive samples, with the Recall indicator reaching above 0.7. As for the division of negative samples, we believe that since the cold start is based on the relocalization results, some originally unknown relationships are predicted to be related, which has an impact on the evaluation of the results.

Table 6. Model performance for 5-fold in CS_train.

|  | AUC | AUPR | Accuracy | F1-score | Precision | Recall | Specificity |
|---|---|---|---|---|---|---|---|
| fold1 | 0.9726 | 0.4967 | 0.9924 | 0.4866 | 0.4427 | 0.5402 | 0.9955 |
| fold2 | 0.9709 | 0.4476 | 0.9913 | 0.4490 | 0.3875 | 0.5336 | 0.9944 |
| fold3 | 0.9688 | 0.4409 | 0.9908 | 0.4377 | 0.3679 | 0.5402 | 0.9938 |
| fold3 | 0.9739 | 0.4629 | 0.9909 | 0.4486 | 0.3751 | 0.5579 | 0.9938 |
| fold5 | 0.9711 | 0.4535 | 0.9910 | 0.4535 | 0.3784 | 0.5658 | 0.9938 |

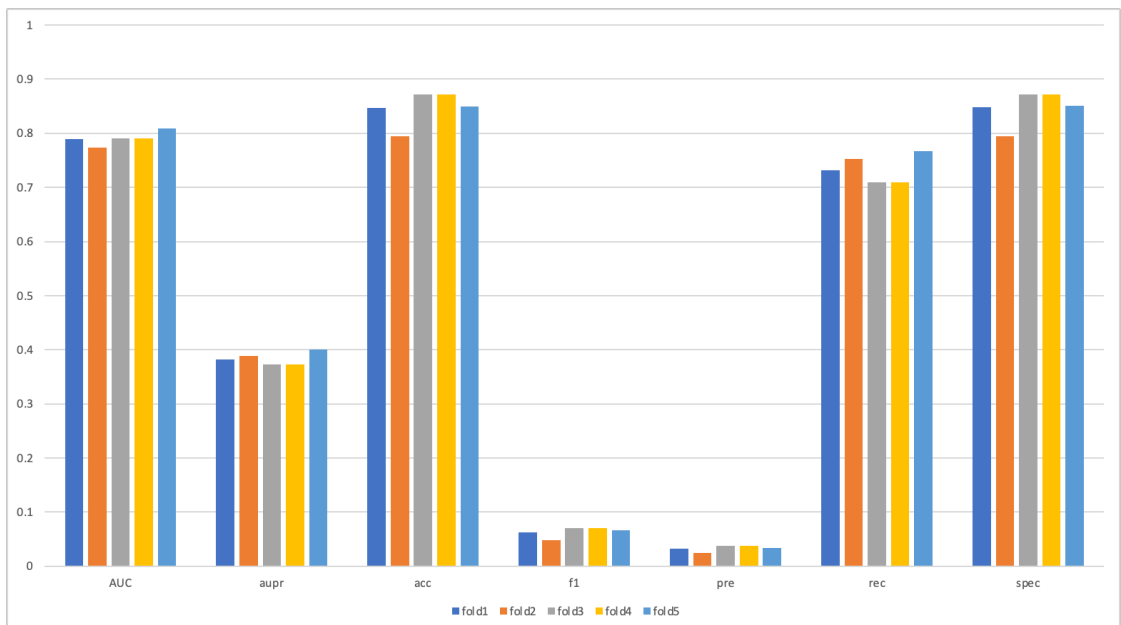

Figure 8. Model cold start task performance in CS_test.

In addition, to verify the significance of the cold start model based on the structural similarity of the drug embedding module, we visualized the results of fold1 (Figure 9.). First, based on the training results of CS_train, we performed t-SNE dimensionality reduction on the vector $R_{train}$ obtained by the drug embedding module and the $R_{test}$ obtained by CS_test in the module, and combined it with KMEANS clustering. The results show that new drugs can indeed be classified into different categories based on structural information. We also sorted the drugs $R_{train}$ and $R_{test}$ according to their distances and displayed them in a heat map. We collected ground label of each new drug and the repositioning results of its highly similar drugs, and calculated the label score between the new drugs of CS_test and drugs of CS_train. And we considered the scores greater than 1.7 are correct classifications. The results show that the overall label score between the new drug and the drug on the left in the heat map is higher than that on the right, indicating that the higher the structural similarity, the more obvious the correlation between the disease and the drug. Therefore, the cold start model based on structural similarity in our task has certain reference value.

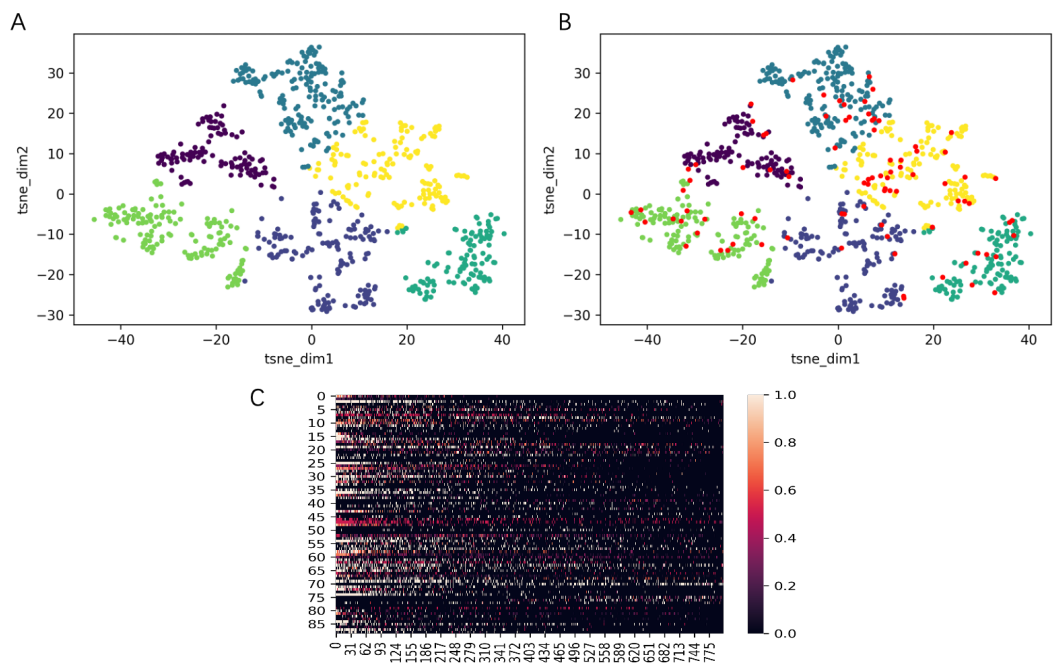

Figure 9. Cold Start Visualization Analysis. A. is the t-SNE and KMeans results of CS_train. B. is the s-SNE result of CS_test. C. is the is the label score sorted based on $R_{train}$ and $R_{test}$ distance.

3.7 SMPR Local Deployment

To further reduce the difficulty of using the model and improve its application potential, we deployed the model locally. Combined with Qt programming tools, SMPR is packaged into a local executable file (.exe). The software is shown in Figure 10. SMPR is a prediction model based on the results of repositioning deep learning. The packaged software provides the user with a SMILE structure input text box, and clicking the "Show Structure" button can display the input structure. When Users click the 'Predict' button, the software will generate a quick_predict.csv result locally based on the input SMILE structure, which includes the scores of input drug and 454 diseases.

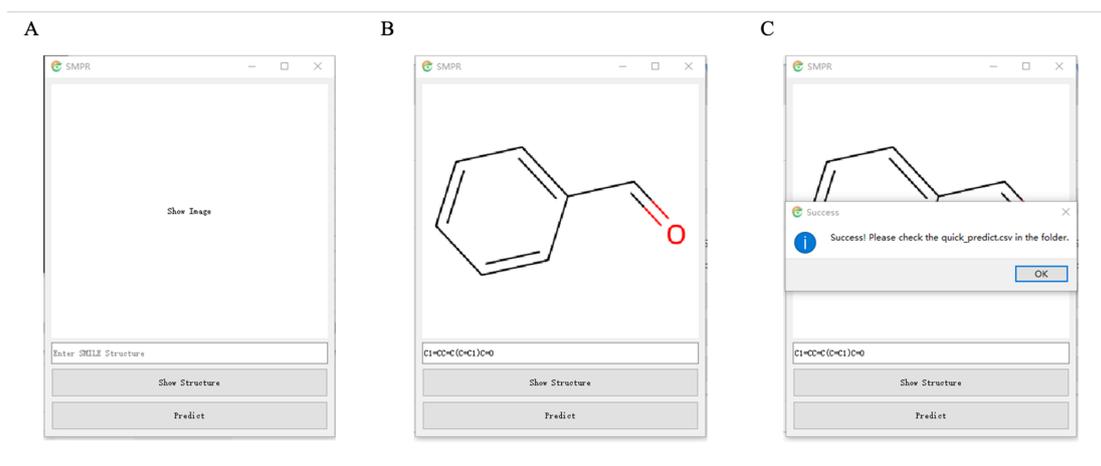

Figure 10. SMPR local deployment. A. The model is packaged into a quick-to-use executable. B. Enter SMILE structure and click 'Show Structure' button, software will show the input drug. C. When click 'Predict' button, a file of predict result will be generated in local.

4. Discussion

The traditional drug discovery work is long and arduous. Especially after experiencing the COVID-19 pandemic, we urgently need precise drug repositioning models to accelerate the process of drug discovery. However, existing models still have shortcomings in the application of structures. The search for potential new target proteins of drugs based on structural similarity is also a major method for drug repositioning tasks. Li et al. [52] screened for novel anti-inflammatory targets of nilotinib through molecular docking. The anticancer effect of benzimidazole was also discovered through a protein stoichiometry method[53]. These cases demonstrate the potential of structures in drug disease repositioning tasks. In addition, existing models also have the problem of poor generalization ability. When facing the problem of drug cold start, most models appear powerless, which makes it difficult to expand the knowledge of repositioning.

We propose a structure-enhanced multimodal prediction model SMRP. This model enhances the focus on drug structure, and the results show that rich structural information has a beneficial effect on improving the accuracy of the model. Compared with existing models, our model has achieved a performance AUC score of 98.7%, AUPR score of 61.06%. And structural information does have a favorable effect on the model's prediction of drug-disease relationships, which is consistent with the fact that drugs with similar structures can enter the same binding pocket. In addition, to verify the robustness of the model, SMPR shows good prediction results in sparse matrix experiments and DC data sets with less modal information. When at least 80% of the drug-disease connection relationships are retained, the model performance does not change significantly. Finally, the literature survey and analysis of two randomly selected drugs, Baclofen (DB00181) and Docetaxel (DB01248), and their effect diseases also verify the application effect of the model, and multiple overlapping pathways are shown between diseases. On the other hand, the model provides a cold start interface to facilitate the task of new drug prediction. Users only need to enter the structural information of the drug, as well as optional supplementary information, to quickly predict. The visualization results show that different drugs can be well classified based on structural information, and drugs with stronger structural correlation also have high correlation in their labels.

However, there are still many parts of the model that are worth improving. With the Alphafold3[54] model winning the Nobel Prize in 2024, researchers are increasingly focusing on the impact of structure. This also means that Alphafold has greatly disclosed unknown protein structure information, which provides the possibility for subsequent models to introduce drug target binding stability and toxicity as auxiliary information. More detailed docking information will also make the practical application of drug repositioning more accurate.

5. Conclusion

In summary, the SMPR model proposed in this study shows excellent performance in drug repositioning tasks and shows stable prediction results in different datasets. In addition, the model also provides a simple interface based on structural similarity for the cold start problem to reduce the difficulty of data processing for actual pharmacologists and accelerate the drug discovery process.

**Abbreviations**
SMPR structure-enhanced multimodal relationship prediction model DA Dataset A
ECFP Extended Connectivity Fingerprint
CS_train Cold start train dataset
CS_test Cold start test dataset DC Dataset C
**Code availability**
SMPR and datasets are provided at https://github.com/dxxxin/SMPR.